\def\year{2021}\relax
\documentclass[letterpaper]{article} 
\usepackage{aaai21}  
\usepackage{times}  
\usepackage{helvet} 
\usepackage{courier}  
\usepackage[hyphens]{url}  
\usepackage{graphicx} 
\usepackage{natbib}  
\usepackage{caption} 
\usepackage{graphics} 
\usepackage{epsfig} 
\usepackage{mathptmx} 
\usepackage{times} 
\usepackage{amsmath} 
\usepackage{amssymb}  
\usepackage{subfig}
\usepackage{caption}
\usepackage{algorithm}
\usepackage[noend]{algpseudocode}
\usepackage[colorlinks=true,linkcolor=black,anchorcolor=black,citecolor=black,filecolor=black,menucolor=black,runcolor=black,urlcolor=black]{hyperref}

\usepackage{subfig}

\usepackage{nameref}

\let\orgautoref\autoref
\providecommand{\Autoref}
{\def\equationautorefname{Equation}%
	\def\figureautorefname{Figure}%
	\def\subfigureautorefname{Figure}%
	\def\Itemautorefname{Item}%
	\def\tableautorefname{Table}%
	\def\exerciseautorefname{Exercise}%
	\def\starexerciseautorefname{Exercise}%
	\def\sectionautorefname{Section}%
	\def\subsectionautorefname{Section}%
	\def\subsubsectionautorefname{Section}%
	\def\chapterautorefname{Section}%
	\def\partautorefname{Part}%
	\orgautoref}

\renewcommand{\autoref}
{\def\equationautorefname{Equation}%
	\def\figureautorefname{Fig.}%
	\def\subfigureautorefname{Fig.}%
	\def\Itemautorefname{item}%
	\def\tableautorefname{Table}%
	\def\exerciseautorefname{Exercise}%
	\def\starexerciseautorefname{Exercise}%
	\def\sectionautorefname{Section}%
	\def\subsectionautorefname{Section}%
	\def\subsubsectionautorefname{Section}%
	\def\chapterautorefname{Section}%
	\def\partautorefname{Part}%
	\orgautoref}

\graphicspath{{figures/}}
\title{Identification of Abnormal States in Videos of Ants Undergoing Social Phase Change}

\author{
    Taeyeong Choi\textsuperscript{\rm 1,2},
    Benjamin Pyenson\textsuperscript{\rm 3},
    Juergen~Liebig\textsuperscript{\rm 3},
	Theodore P.~Pavlic\textsuperscript{\rm 2,3,4} \\
}
\affiliations{%
	\textsuperscript{\rm 1} Lincoln Institute for Agri‐food Technology,
	University of Lincoln, Riseholme Park, Lincoln, UK \\
	\textsuperscript{\rm 2} School of Computing,
	Informatics, and Decision Systems Engineering,
	Arizona State University, Tempe, AZ, USA \\
	\textsuperscript{\rm 3} School of Life Sciences,
	Arizona State University, Tempe, AZ, USA \\
	\textsuperscript{\rm 4} School of Sustainability,
	Arizona State University, Tempe, AZ, USA \\
	tchoi@lincoln.ac.uk, \{bpyenson, jliebig, tpavlic\}@asu.edu

}
\begin{document}

\maketitle

\begin{abstract}
Biology is both an important application area and a source of motivation
for development of advanced machine learning techniques. Although
much attention has been paid to large and complex data sets
resulting from high-throughput sequencing, advances in
high-quality video recording technology have begun to generate
similarly rich data sets requiring sophisticated techniques from both
computer vision and time-series analysis. Moreover, just as studying
gene expression patterns in one organism can reveal general principles
that apply to other organisms, the study of complex social
interactions in an experimentally tractable model system, such as a
laboratory ant colony, can provide general principles about the dynamics
of other social groups. Here, we focus on one such example
from the study of reproductive regulation in small laboratory colonies
of more than 50~\emph{Harpegnathos} ants. These ants can be
artificially induced to begin a $\sim$20~day process of hierarchy
reformation. Although the conclusion of this process is conspicuous to a
human observer, it remains unclear which behaviors during the
transient period are contributing to the process. To address this issue, we
explore the potential application of One-class Classification~(OC) to
the detection of \emph{abnormal} states in ant colonies for which
behavioral data is only available for the \emph{normal} societal
conditions during training. Specifically, we build upon the Deep Support
Vector Data Description~(DSVDD) and introduce the Inner-Outlier
Generator~(\mbox{IO-GEN}) that synthesizes fake ``\emph{inner outlier}''
observations during training that are near the center of the DSVDD data
description. We show that \mbox{IO-GEN} increases the reliability of the final
OC~classifier relative to other DSVDD baselines. This method can be used
to screen video frames for which additional human observation is needed.
Although we focus on an application with laboratory colonies of social
insects, this approach may be applied to video data from other social
systems to either better understand the causal factors behind social
phase transitions or even to predict the onset of future transitions.
\end{abstract}

\section{Introduction}
\label{sec:intro}

\noindent In natural social systems, complex interactions among large
numbers of individuals can give rise to phenomena such as ``collective
minds''~\citep{C07} and, as in colonies of ants, even ``superorganisms''
where the collective can be described as a single monolithic entity.
Some ant species have colonies sufficiently small to be observed 
in their entirety with state-of-the-art video-recording technologies 
and sufficiently large to have rich, multi-scale behaviors. Whereas 
the dynamical processes underlying the non-trivial interactions between ants
are cryptic to human observers, there is potential for
machine-learning and artificial-intelligence techniques to identify social
interaction patterns that warrant further study. For example, in species
of ants where new reproductive individuals emerge after
the previous reproductive dies naturally or is artificially 
removed~\citep{HHP94, SPSHPL16}, machine learning could in principle 
help to identify abnormal patterns that only occur during this conflict 
resolution. However, such a behavioral
classifier for underappreciated \emph{abnormal} patterns would
necessarily be limited to training data from videos of behaviors under
known \emph{normal} conditions.

\begin{figure}[t]\centering
    \includegraphics[width=0.8\columnwidth]{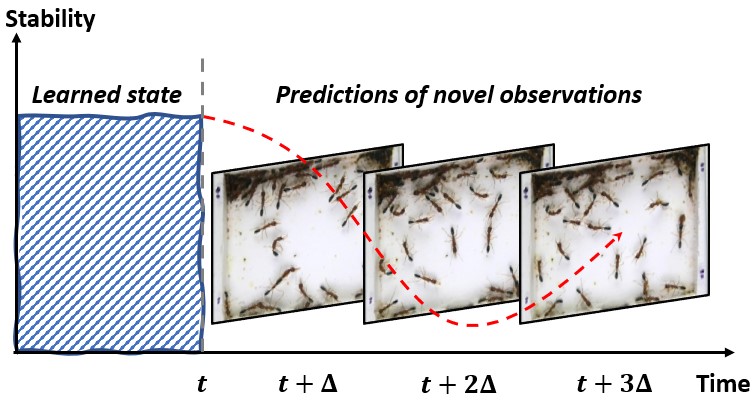}
	\caption{Proposed scenario in which observational data from 
	an ant colony are accessible during its stable state, 
	while the trained predictive model is to classify behaviors 
	from unseen, unstable states.}
	\label{fig:concept}
\end{figure}
Here, we propose an alternative application of
One-class Classification~(OC) to solve
the \emph{abnormal state detection problem} for video data of social
systems in transition from disordered to ordered states.
In these systems, behavioral data for training
the classifier is only available for the system's typical state,
but the classifier must be able to classify abnormal samples presented
to it after training.
For OC problems, Support-Vector-Machine--inspired approaches have widely
been used with the combination of autoencoders, which can learn key
features in an unsupervised manner while significantly reducing the
number of dimensions of original input~\citep{XRYSS15, RGLL20}. One of
the most successful algorithms of this sort is Deep Support Vector Data
Description~(DSVDD)~\citep{RVGDSBMK18}, which learns a hyperspheric
feature space where samples of an available class lie densely around
a central point~$\vec{c}$ so that the distance from it is used as the
indicator of novelty during test. We argue that the DSVDD distancing approach
may oversimplify useful relationships among features for OC
especially in high-dimensional spaces. Thus, we propose a
generative module, \emph{Inner-Outlier Generator~(\mbox{IO-GEN})}, to replace
the heuristic reference~$\vec{c}$ with synthesized ``\emph{inner
outlier}'' observations of an imaginary social-system state so that a
separate classifier on DSVDD can use both the hyperspheric structure of
data description and high-dimensional feature representation to learn
behavioral abnormality.

We apply this \mbox{IO-GEN} approach to the analysis of a group of Indian 
jumping ants~(\emph{Harpegnathos saltator}), which typically show a 
\emph{stable}~(normal) state with mainly peaceful interactions. 
When the reproductive ants die or are artificially removed, 
the colony moves into a transient \emph{unstable}~(abnormal) state during 
which members go through a process that comes to consensus on a subset of 
workers that become the new reproductive individuals leading to a 
new stable state~\citep{SPSHPL16}.
There are conspicuous behavioral interactions that only occur during 
the unstable colony state, but a detailed understanding of how this
transitory period is resolved remains elusive. 
We created a dataset for analysis by
extracting optical flows from a colony of over $50$~\emph{H.~saltator}
ants to record their behavioral data for $20$~days in a lab setting
where the colony was artificially triggered to induce
``stable--unstable--stable'' colony state transitions. We then developed
an approach following the simplified diagram in \autoref{fig:concept}.
Video data of a particular stable colony is used for normal-class
training, and our proposed model then later assesses whether a focal
colony is stable or unstable based on a short sequence of new optimal
flow inputs.

\section{Related Work}
\label{sec:related_work}

\subsection{Behavioral Cues for Inferring Collective States}
\label{sec:behavioral_cues_to_differentiate_collective_states}

Inference and prediction of current and future collective states is
potentially useful in a number of applications. In human crowds,
intelligent surveillance cameras can detect abnormal collective
states~(e.g., conditions consistent with group-level panic or rioting
behavior)~\citep{MOS09} so that authorities can prioritize surveillance
resources and execute proactive mitigation strategies. Alternatively, an
individual robot in a multi-robot system can use local information of
the pose of nearby robots to infer the large-scale formation of its team
and then alter its own trajectory to more effectively achieve a
group-level response to a stimulus encountered at distal ends of the
team~\citep{CPR17, CKP20}.

In behavioral ecology, however, modeling efforts have been focused
either on the coarse-grained collective scale or the fine-grained
individual scale but rarely the connection between the two. For example,
many mathematical and statistical models have been developed for
understanding the evolution of group-level states and how they adapt to
changes in the environment~\citep{CKJRF02, RLPKCG15, SPSHPL16, PMSF02,
BSCHDMS06}. These approaches provide insights into the overall function
of collective states but do not provide so much insight into how to map
observations of an individual to the collective context of that
individual. On the other end of the spectrum, more recent deep learning
approaches for image segmentation or object detection have been tuned to
track individual animals from video frames~\citep{BHMS18, NMCPBM19}.
These efforts are focused on accelerating data acquisition for existing
statistical pipelines that human researchers employ and not on making
automated inferences across the individual--group scales. \Citet{CPM19}
used an unsupervised learning framework to discover latent states in
\emph{Drosophila melanogaster} flies during courtship, but the inference
scale was only limited to the group of two engaged flies while our work
deals with much larger social groups~(an entire animal society).

\subsection{One-class Classification~(OC) for Visual Data}
\label{sec:occ}

\begin{figure}\centering
	\includegraphics[width=.7\linewidth]{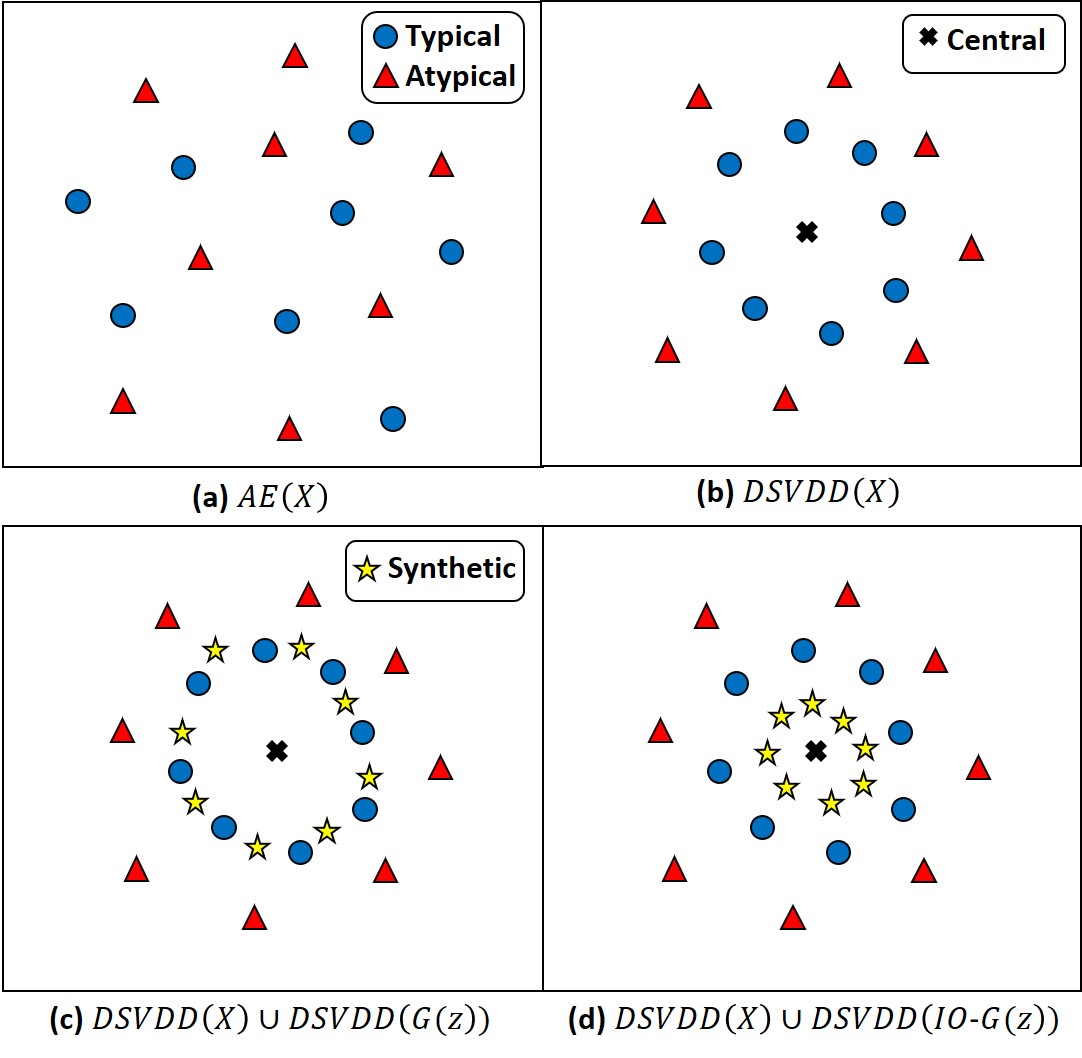}
	\caption{
        Conceptual feature formations in different methods on
        two-dimensional planes assuming only typical examples have been
        used for training each model.
        (a)~The encoding from autoencoder cannot ensure a clear
        separation of atypical samples.
        (b)~DSVDD forms the space in which data of seen class surround a
        central point~$\vec{c}$ more closely than unseen examples.
        (c)~Generators in GANs learn to produce typical properties used
        for training.
        (d)~\mbox{IO-GEN} synthesizes inner outliers much more densely to later
        substitute for~$\vec{c}$.
	}
	\label{fig:feature_space}
\end{figure}

Classical OC methods, such as One-class Support Vector
Machine~(OC-SVM)~\citep{SPSSW01} and Support Vector Data
Description~(SVDD)~\citep{TD04}, either use a hyperplane or a
hypersphere tightly bounding the known-class data for separation.
Recently, these methods have been augmented with autoencoders that
highly reduce input dimensions of graphical data without
supervision~\citep{XRYSS15, RGLL20}. As an extension, DSVDD is designed
to optimize the objective of SVDD in an end-to-end deep neural network
pipeline~\citep{RVGDSBMK18}. In particular, the autoencoder's encoder is
fine-tuned to generate a feature space in which the in-class
samples lie densely close to a pre-defined central vector~$\vec{c}$,
while the out-of-class samples are sparsely away from
them~(\autoref{fig:feature_space}b).
Although DSVDD showed competitive performance with several benchmark
datasets, we argue that its distancing scheme~-- simply using the
distance to $\vec{c}$~-- is not sufficiently rich to distinguish novel
samples, and we combine DSVDD with a generative approach to improve OC
for this case.

Generative Adversarial Networks~(GANs)~\citep{GPMXWOCB14} can also be
used in OC by synthesizing fake outcomes consistent with typical samples
that are then used in training to improve the generalizability of the
OC~\citep{SKFA18, YCR20, PNX19}. However, these approaches adopt the
conventional min--max scheme of GANs to closely emulate the data
distribution of the available class~(\autoref{fig:feature_space}c)
although the ultimate goal is to identify novel samples from a different
distribution. Instead, our \mbox{IO-GEN} generates fake outcomes even closer to
the idealized central vector $\vec{c}$~(\autoref{fig:feature_space}d).
These more prototypical samples allow the subsequent classifier to learn
sharper discrimination in the DSVDD feature space between normal and
abnormal samples.




\section{Background: \emph{Harpegnathos saltator}}
\label{sec:background}
%
%

A key characteristic of an ant colony is a reproductive division of 
labor between reproductive and non-reproductive individuals. 
In most ant colonies, a single ``queen'' is chiefly responsible for 
laying eggs that develop into non-reproductive ``workers'' that 
are responsible for caring for the next generation of eggs. 
Because workers typically cannot produce
new workers themselves, the death of a queen usually means the
expiration of the colony shortly after. 
In contrast, in the case of \emph{H.~saltator}, workers
have the ability to produce all types of individuals including new
workers,
but they do not lay eggs while another
reproductive is present. However, when there is no living reproductive
in a colony, mated workers engage in a hierarchy reformation process 
lasting several weeks that terminates when several mated workers activate
their ovaries and begin to produce eggs~\citep{LPH99, SPSHPL16}. The
reproductive ascendance of these workers, known as 
``gamergates''~\citep{PC85}, brings the colony back to a typical state. 
When those gamergates die or are removed, the instability will begin 
again~\citep{LPH99, SPSHPL16}. 
During the unstable transient state, workers perform conspicuous aggressive
behaviors such as \emph{dueling}, \emph{dominance biting}, and 
\emph{policing}~\citep{SPSHPL16}. 
Although these behaviors are clear signs that the colony is in 
this transient state, the precise combination of events that 
leads to a colony-level stable state remains unclear.



\section{Dataset from Colonies in Transition}
\label{sec:dataset}

Deep-learning methods for image and video data have shown that optical
flows can effectively complement classical RGB data in learning because
they can extract transient behavioral characteristics (e.g., shooting)
while RGB data largely provides an understanding of scenic context with
visible objects (e.g., bow and arrows)~\citep{SZ14}. Because our
framework only expects ants and crickets they feed upon in
the scenes, we only use optical flows in our datasets so that learning
will be based solely on behavioral flows, which is similar to
the human-crowd behavior classification by \citet{MOS09}.

We used a colony of $59$~\emph{H.~saltator} ants including $4$~gamergates 
in a plastic nest covered by a transparent glass. Although an overhead camera 
filmed the nest, not all ants necessarily appear in all scenes because some may 
move to an off-camera foraging chamber through a tunnel on the left side 
of the nest. 
Videos were recorded for $20$~days, denoted as D-2, D-1, D+1,
\dots, D+18, where D-0 represents the instant of removal of the previously  
identified gamergates between days~$2$ and~$3$ to artificially trigger
the transient state of the colony. From D+1, we observed frequent
\emph{dueling} and \emph{dominance biting} until the aggressiveness
almost disappeared on the last several days across the group. By
performing downsampling techniques, $m$~sequential optical flows were
sampled every $2$~minutes, and, for each flow, a pair of horizontal and
vertical motional representations in the spatial resolution of $64
\times 64$ were extracted from two consecutive frames with an interval
of $0.5$~seconds. The code provided by \citet{WXWQLTV16} was used to
acquire $1,333m$~stable-class and $11,984m$~unstable-class optical flows
in total. Three unique splits of stable class were prepared to obtain
the average performance of three separate models, as $80\%$ and $20\%$
were used for training and test, respectively in each split, while the
unstable samples were included only in the test sets. All the data and 
split information are accessible
online\footnote{\url{https://github.com/ctyeong/OpticalFlows_HsAnts}}.

\section{Methodology}
\label{sec:method}

\begin{figure}
	\centering
		\includegraphics[width=0.45\linewidth]{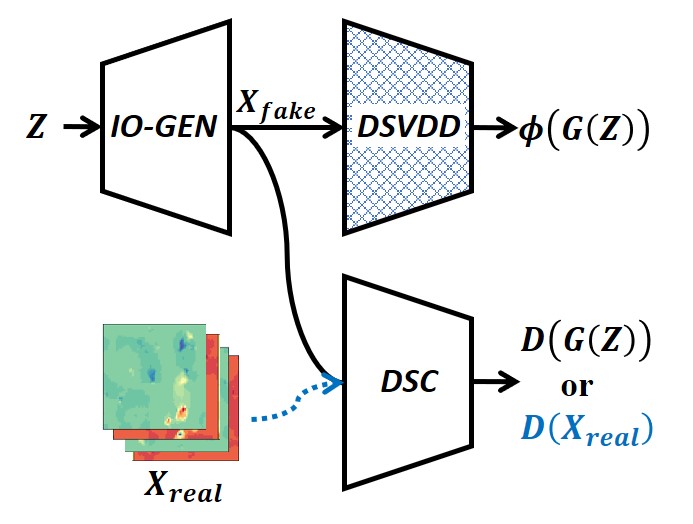}%
		\includegraphics[width=0.45\columnwidth]{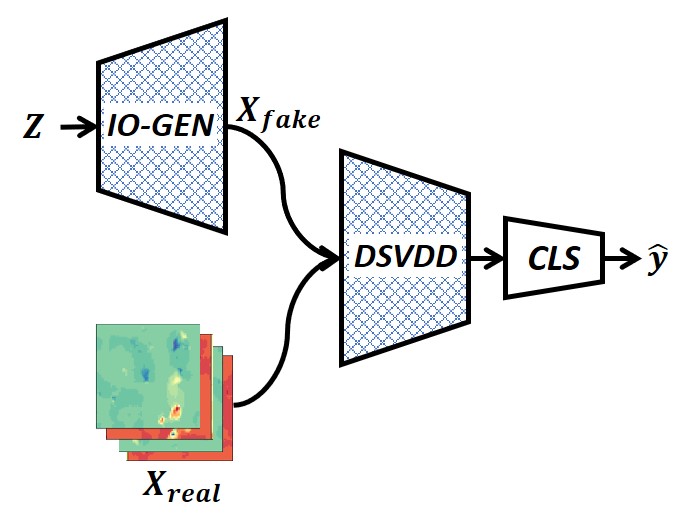}%
    %
    \caption{Training pipelines for \mbox{IO-GEN}~(left) and
        Classifier~(right).
        \mbox{IO-GEN} must meet two objectives simultaneously, one with 
		the pre-trained DSVDD and one with the discriminator. 
		Classifier learns binary classification on the data description 
		that the DSVDD offers. 
		Patterned components represent their parameters fixed during 
		the training phase.  
        }
    \label{fig:structure}
\end{figure}
%

\subsection{DSVDD}
\label{sec:dsvdd}

DSVDD in our framework follows its original design
from~\citet{RVGDSBMK18}. It is built from the encoder part~$\phi$ of a
pre-trained autoencoder that is used to learn a feature
space~$\mathcal{F}$ in which the samples of known class have a lower
average distance to a central vector~$\vec{c}$ than those of novel
class. Specifically, we adopt One-class DSVDD, which minimizes the
objective:
\begin{equation*}
	\label{eq:dsvdd_obj}
		\min_{W} \frac{1}{n} \sum_{i=1}^{n} ||\phi(x_{i}; W) - \vec{c}||^{2} \
		+ \frac{\lambda}{2} \sum_{l=1}^{L} || W^l ||_{F}^{2}
\end{equation*}
where $|| \cdot ||_{F}$ is the Frobenius norm. The first term is closing
the distance between $\vec{c}$ and the feature representation of each
sample~$x_i$ in encoder~$\phi$ parameterized by $W$, and the second term
is a weight decay regularizer for $L$~layers with $\lambda > 0$. In the
original method of DSVDD, the trained parameters $W=W^*$ are used to
generate a distance:
\begin{equation*}
    \label{eq:dsvdd_score}
    s(x) = ||\phi(x;W^{*}) - \vec{c}||^2
\end{equation*}
that is a proxy for how atypical a sample $x$ is. For some threshold
$\tau > 0$, $s(x) > \tau$ classifies $x$ as atypical. Our method,
however, substitutes the distancing heuristic~$s(x)$ by \mbox{IO-GEN} and
Classifier, described below, that we argue better utilize key features
in the normal set in order to discriminate abnormal data after training.

\subsection{IO-GEN}
\label{sec:io_gen}

As shown in~\autoref{fig:structure}, \mbox{IO-GEN}~$G$ is designed to operate
with both the pre-trained DSVDD~$\phi$ and a discriminator network~$D$,
as a generative model of optical flows. With the discriminator, an
adversarial learning is performed following the standard objective:
\begin{equation*}
    \label{eq:obj1}
    \min_{G} \max_{D}
    \Big( \mathbb{E}_{z \sim \mathit{N_{\sigma}}} [\log(1 - D(G(z)))]
    + \mathbb{E}_{x \sim p} [\log(D(x))] \Big)
\end{equation*}
where $\mathit{N_{\sigma}}$ is the zero-mean normal distribution with
standard deviation $\sigma$, and $p$ is the probability distribution of
real optical-flow data. Due to the first term, the outcomes from
\mbox{IO-GEN} are adjusted to appear sufficiently realistic to deceive the
discriminator. Also, the DSVDD is used to force the learned synthetic 
data to be inner outliers close to $\vec{c}$ in~$\mathcal{F}$, while the
parameters of itself are not updated. In particular, we use the
feature-matching technique proposed by \citet{SGZCRC16}, which
incorporates the minimization:
\begin{equation*}
	\label{eq:obj2}
	\min_{G}
	|| \mathbb{E}_{z \sim \mathit{N_{\sigma}}} [\phi(G(z))] 
	- \vec{c} ||^{2}_{2}
\end{equation*}
%
%
The composite loss function for \mbox{IO-GEN} is:
\begin{multline*}
\label{eq:loss}
	\mathit{L}_{G} = 
	\mathbb{E}_{z \sim \mathit{N_{\sigma}}} [\log(1 - D(G(z)))] \\
	+ \lambda \Big( || \mathbb{E}_{z \sim \mathit{N_{\sigma}}} \
	[\phi(G(z))] - \vec{c} ||^{2}_{2} \Big)
\end{multline*}
where hyperparameter $\lambda > 0$ determines the relative
weights between the two terms to minimize. In other words, \mbox{IO-GEN} is
trained to produce ant behavioral flows that look real and
also feature the closest proximity to $\vec{c}$ in $\mathcal{F}$ of
DSVDD.

\subsection{Classifier}
\label{sec:classifier}

Classifier utilizes real data from stable colony states as well as the
generated \mbox{IO-GEN} inner outliers to learn to predict the likelihood of
unstable behaviors on given $m$~instant frame images as in general
binary classifiers. We introduce a novel strategy, \emph{label switch}, during
training where the real stable samples are labelled as
``unstable''~(atypical) and the synthetic ones are labeled as
``stable''~(typical). This technique leads the Classifier to eventually
make \mbox{low-,} \mbox{mid-,} and high-range likelihood predictions for synthetic,
stable, and unstable data, respectively, as though the augmented state
of inner outliers was the ``most stable'' state . That is, Classifier
offers likelihood outcomes somewhat consistent with the class
distribution around $\vec{c}$ allowing for a clear separation between
real stable and unstable data, which will be demonstrated in
the following sections.

\subsection{Model Structures \& Relevant Parameters}
\label{sec:model_structures}

A Deep Convolutional Autoencoder~(DCAE) is used as the backbone of DSVDD
and \mbox{IO-GEN} once it has been trained with the data of stable class to
minimize the reconstruction error in the Mean Squared Error~(MSE)
between the input encoded and the decoded output. Per input, 
$m$~optical flows are all stacked one another to constitute an
input $x \in \mathbb{R}^{64 \times 64 \times 2m}$ after normalization to
range in $[-1, 1]$. In the encoder, three convolutional layers with $32,
64,$ and $128$ 2D kernels are employed in series as each kernel is of $3
\times 3$ size. Also, every output is followed by a \emph{ReLU}
activation and 2D maxpooling. The decoder has the reversed architecture
of the encoder with two modifications: 2D upsampling instead of
maxpooling and an added output layer with $2m$~kernels and
\emph{hyperbolic-tangent}~($\tanh$) activation.
An additional $32$~convolutional kernels are placed as the
encoder--decoder bottleneck to obtain a compact
encoding scheme~$\vec{v} \in \mathbb{R}^{1 \times 2048}$ when flattened.
DSVDD takes advantage of the pre-trained encoder to reshape the space
of~$\vec{v}$ by learning data description~$\mathcal{F}$ as the reference
vector~$\vec{c}$ is the mean of available encoded
samples according to~\citet{RVGDSBMK18}.

\mbox{IO-GEN} essentially employs a fully connected layer with the \emph{ReLU}
activation that takes a noisy vector $\vec{z} \in \mathbb{R}^{1 \times
100}$ as input. It is then connected to a replica of the pre-trained decoder
so that realistic synthesis can be learned faster from the prior
knowledge of reconstruction. The discriminator network builds an extra
fully connected layer with a \emph{sigmoid} activation on top of
encoder, but its weights are all reinitialized because otherwise it
appears to easily overwhelm \mbox{IO-GEN} in performance causing unstable
adversarial training. Also, $\lambda=10$ was empirically found most
effective to minimize $\mathit{L_G}$.

Because Classifier comes after DSVDD, it has an independent architecture
in which five convolutional layers learn $8, 16, 24, 48, \text{ and }
48$ one-dimensional kernels, respectively. Each layer has a
\emph{LeakyReLU} activation~($\alpha=0.3$) and 1D average pooling, and
lastly, a fully connected layer is deployed to provide a predicted
likelihood of unstable state via a \emph{sigmoid} function.
All codes are also available 
online\footnote{\url{https://github.com/ctyeong/IO-GEN}}.

\section{Experiments}
\label{sec:experiments}


\subsection{Observations of Optical Flow Weights}
\label{sec:obs_optical_flow_weights}

\begin{figure}[t]
	\centering
	\includegraphics[width=.7\columnwidth]{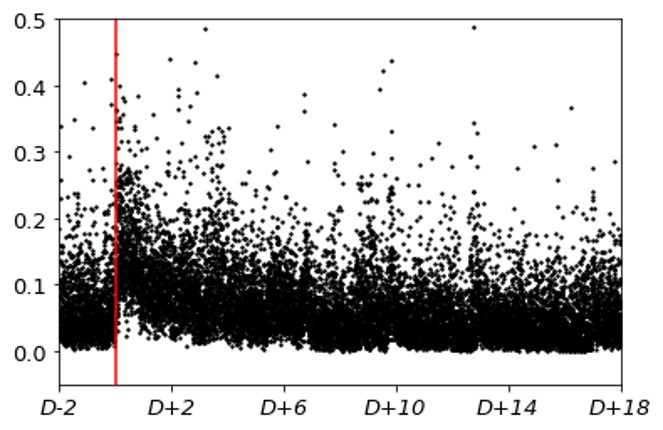}
	\caption{
		Optical flow weights each sampled at the interval of 
		over $2$~minutes for $20$~days. 
		The horizontal axis shows the days of observation 
		with the red separator
		when the \emph{gamergates} were intentionally moved.
		The vertical axis indicates the weight levels, each 
		standardized in $[0, 1]$ by the global max and 
		min, omitting extreme outliers for clarity.
	}
	\label{fig:ofw}
\end{figure}
We first attempt to make use of the optical flow dataset to discover
insightful motional patterns without any complex model. As with
\citet{MCFT13}, we compute an optical flow weight~$w_i$ for each
frame~$i$ by averaging the magnitudes of flow vectors at all locations.
\Autoref{fig:ofw} displays the obtained weight signal in time as
$m=1$~optical flow frame is considered for each sampling interval.

For the first two days, the weights generally stay in a narrow range
implying some behavioral regularity maintained among ants. Then there is
a noticeable increase in weights starting at D+1 just after the gamergates
are removed because the removal triggered a social
tournament involving frequent aggressive behaviors. As the
transient period evolves, the magnitude continues to decrease and
finally recovers the original extent roughly on D+10 even though
several ants continue to present hostile interactions at that time.

Consequently, a simple model might be built that uses the overall rise
of flow weight as the only feature to distinguish the unstable colony
from the stable especially at early development of the unstable state.
However, following our results, we will provide concrete examples that
show the limitations of such a design and the need of more complex
models for reliable predictions.

\subsection{Model Evaluation}
\label{sec:model_evaluation}

Here, we demonstrate the OC performance of our proposed method. We first
describe an ablation study to find the best number of optical-flow
frames per input. Next, baselines used for comparison are introduced
that will help us explore our method's overall reliability and
prediction robustness in various time windows during colonial
stabilization.

As in previous works~\citep{RVGDSBMK18}, the Area Under the Curve~(AUC) 
of the Receiver Operating Characteristics~(ROC) are measured for 
each model to reflect the separability between classes. Moreover, 
the average over three splits is reported with the standard 
deviation when needed.

\subsubsection{Ablation Study:}
\label{sec:ablation_study}
%
Results from tests with $m \in \{1, 2, 4\}$ are shown in
\autoref{table:ablation_study}.
\setlength{\tabcolsep}{0.5em} 
{\renewcommand{\arraystretch}{1.2}
	\begin{table}
		\centering
		\begin{tabular}{|c|c|c|c|} 
			\hline
			$m$  &  1 & 2 & 4  \\ \hline\hline
			AUC & $0.760~\pm 0.016$ & $0.786~\pm 0.009$ & $0.787~\pm 0.008$ 
			\\ \hline
		\end{tabular}
		\caption{Average performance when the number of optical flow 
		image frames per input is set to $1, 2,$ or $4$.}
		\label{table:ablation_study}
	\end{table}
}%
There was an improvement as $m$ increased from $1$ to $2$,
while doubling it to $4$ did not offer any benefit. The result may
indicate that the observation of one more second does not add
significantly more information. Learning \mbox{IO-GEN} could also be more
challenging as it is asked to generate longer motional sequences.  Thus,
$m$ is set to $2$ hereafter considering both efficiency and
effectiveness of our model.


\subsubsection{Baselines:}
\label{sec:baselines}

\textbf{OFW} uses the temporal optical flow weights to set the best
threshold to report the best classification result. \textbf{DCAE} is a
similar threshold-based method relying on the reconstruction error as
the feature of novelty~\citep{KWWBKA19}. \textbf{OC-SVM}~\citep{SPSSW01}
takes the encoder of DCAE to build the One-class SVM on it providing the
performance with the best $\nu$ parameter. While \textbf{DSVDD} here is
designed similarly to the description by~\citet{RVGDSBMK18}, the
adjustments in our implementation are described
in~\nameref{sec:model_structures} above. \textbf{GEN} and
\textbf{\mbox{N-GEN}}
are generative models to train a separate classifier as our method. GEN
is, however, a standard generative model adopting the feature matching
technique in the discriminator network instead without the intervention
of DSVDD. \mbox{N-GEN} replaces $\phi(G(z))$ with arbitrary noisy
data~$\vec{v}'\in\mathbb{R}^{1\times2048}$ where each element
of~$\vec{v}'$ is drawn from $\mathit{N}(0, \alpha)$ where $\alpha$ is
the global variation of $\vec{v}\sim\phi(G(z))$.

\subsubsection{Overall Performance:}
\label{sec:overall_performance}

\setlength{\tabcolsep}{0.5em} 
{\renewcommand{\arraystretch}{1.2}
	\begin{table}[t]
		\centering
		\begin{tabular}{|p{40mm}c|} 
			\hline
			METHOD & AUC \\ \hline\hline
			OFW & $0.506$ \\ 
			DCAE & $0.506~\pm 0.002$ \\ 
			OC-SVM & $0.523~\pm 0.004$ \\ 
			DSVDD & $0.762~\pm 0.013$ \\ 
			GEN & $0.587~\pm 0.032$ \\ 
			N-GEN & $0.699~\pm 0.006$ \\ \hline\hline
			IO-GEN & $\mathbf{0.786~\pm 0.009}$ \\ \hline
		\end{tabular}
		\caption{Average AUC of tested models with the standard deviation 
		as all $18$-day unstable observations are considered.}
		\label{table:comparison}
	\end{table}
}
\Autoref{table:comparison} helps estimate overall reliability of each
model for the image inputs that can be captured at an arbitrary timing
since all samples from unstable colony were included for test. OFW and
DCAE suggest the limitation of only relying on thresholding a simplistic
one-dimensional signal. In particular, the low accuracy of DCAE implies
that precise reconstruction is achieved also for the motions from 
unseen, unstable colony. 
Similarly, the OC-SVM can utilize only little benefit from the
encoding capability. On the other hand, DSVDD leads at least $45\%$
increase of AUC score simply fine-tuning the encoder part of DCAE
because unstable examples are more easily distinguished in the newly
learned hyperspheric data description. In addition, our model brings
about a further improvement proving that utilizing a subsequent
classifier with synthetic examples can be more effective than the
distancing heuristic in DSVDD to make full use of multi-dimensional
relationships among features. Nevertheless, GEN and \mbox{N-GEN} provide $25\%$
and $11\%$ poorer performance than ours although both also use synthetic
data to train a classifier. \mbox{N-GEN} actually performs better than GEN
implying that the prior knowledge on data description is useful for
effective data synthesis. Still, its insufficient reliability emphasizes
the importance of realism in generated datasets as well.

\subsubsection{Detection in Different Developmental Phases:}
\label{sec:detection_in_different_developmental_phases}

\begin{figure}[t]
	\centering
	\includegraphics[width=.75\columnwidth]{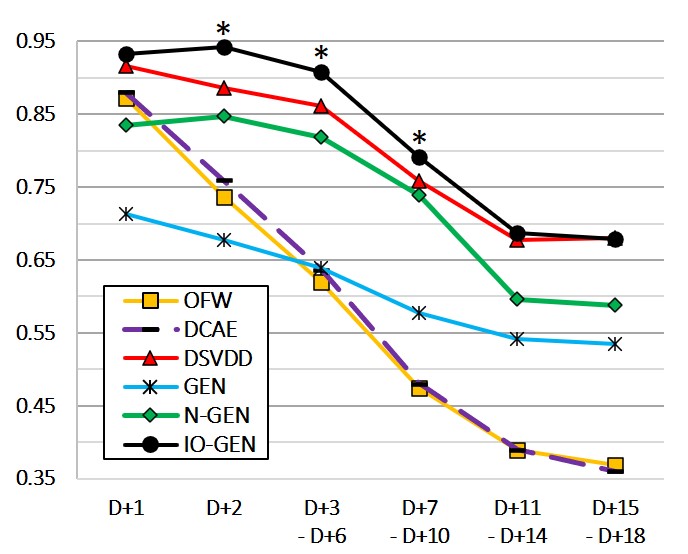}
	\caption{
		Average AUC changes for predictions within different temporal
		windows.
        Asterisks (*) mark statistically significant improvement over
		DSVDD~($p<.05$).
	}
	\label{fig:auc_less_data}
\end{figure}
\begin{figure}\centering
       \includegraphics[width=.75\linewidth]{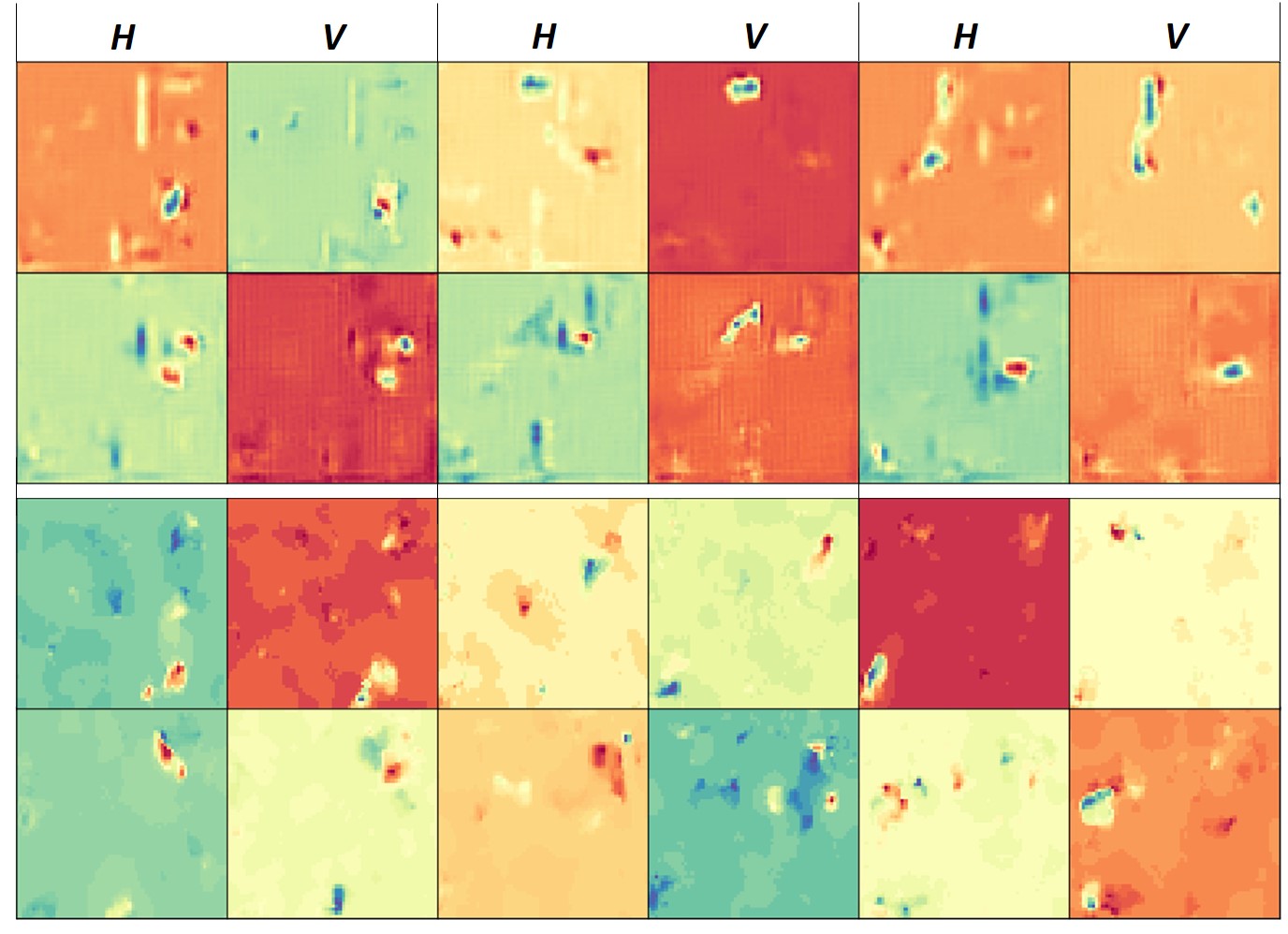}
       \caption{Optical flow examples:
           (top two rows)~Six synthesized pairs from \mbox{IO-GEN};
           (bottom two rows)~Six real examples.
           Each (H-V) pair show horizontal and vertical motions,
           respectively, for which pixels are normalized in each image.
       }
       \label{fig:real_fake_flows}
\end{figure}
\Autoref{fig:auc_less_data} displays the performance variation of each
model as the tested data from the unstable colony are confined in 
various temporal windows. 
Consistent with \autoref{fig:ofw}, the prediction performance 
generally degrades for later temporal bins because the ant colony is 
more stabilized. Our framework still indicates the top performance in
almost any phase especially presenting the highest margins from DSVDD 
in a highly ambiguous time period between D+2 and D+10, in which the 
proportion of stable observations dramatically increased. 
As expected from~\autoref{fig:ofw}, OFW and DCAE highly depend on the
timing of application because their scores are close to that of DSVDD
early while lower even than $0.5$ after D+6. 
If the initial social transition is less conspicuous, possibly due to 
a smaller population, these models may perform poorly because of less 
intense competition caused. 
Moreover, the results from GEN and \mbox{N-GEN}
reemphasize the insufficiency of solely relying on realism or 
spatial characteristics of produced features when training generative 
models.
In particular, as illustrated in~\autoref{fig:feature_space}, 
GEN produces fake samples that closely resemble the ones of stable state,
and so the biased classifier leads to the worst performance in
the early stages~($\sim$D+2) when colonial instability was highest.

\subsection{Model Properties}
\label{sec:model_properties}


\begin{figure}
	\centering
    \includegraphics[trim={0pt 30pt 0pt 0pt},clip,width=.98\columnwidth]{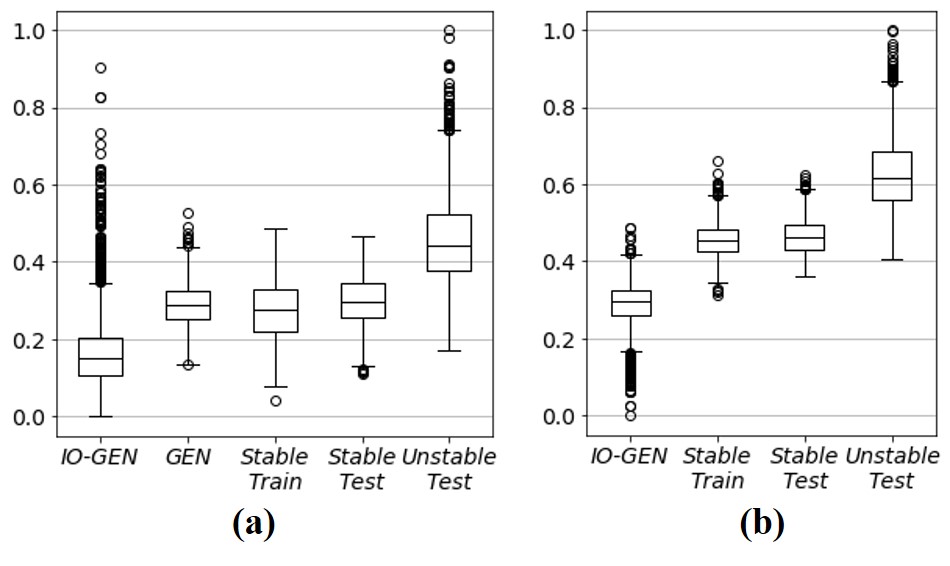}
	\caption{For different types of data:
        On left, normalized Euclidean distances to $\vec{c}$
		in feature description~$\mathcal{F}$ of DSVDD.
        On right, predicted likelihoods from Classifier.
	}
	\label{fig:dist_to_c}
\end{figure}
\Autoref{fig:real_fake_flows} compares synthetic optical flows from
\mbox{IO-GEN} to real optical flows; the generated optical flows are
visually similar to real flows. Furthermore, \autoref{fig:dist_to_c}a
illustrates that the lowest distance distribution to $\vec{c}$ is
measured with \mbox{IO-GEN}, as designed, whereas GEN behaves similarly to the
stable dataset. \Autoref{fig:dist_to_c}b finally shows the predictive
outcomes of Classifier, which are likelihoods of unstable state. With
the \emph{label switch}, the confidence becomes positively correlated
with the distance to $\vec{c}$ viewing inner outliers as samples from
the most stable colony. Clear differences between classes imply that
learned knowledge to discriminate stable and more-stable states in DSVDD
can be transferred for classification of another pair as stable or
unstable.




\section{Conclusion}
\label{sec:conclusion}

We have introduced a novel generative model~\mbox{IO-GEN} that can utilize
a pre-trained DSVDD and a separate classifier to successfully
solve the OC problem. Our framework has been applied to $20$-day
video data from an entire society of $59$~\emph{H.~saltator} 
ants to identify a colony's stable or unstable state only from a 
$1$-second motional sequence. Experiments have shown that the 
classifier trained with the synthetic data from \mbox{IO-GEN} outperforms 
other state-of-the-art baselines at any temporal phase during social 
stabilization.

Our future directions include a graphical user interface for
this method that acts as a tool for biologists that can propose 
frames or individuals~(regions of interest) implicated 
as being crucial in the evolution of social state. 
To implement this, an additional module can be built to monitor and 
visualize the levels of gradient passing from spatio-temporal behavioral 
features to the final decision output~\citep{Choi2020PhDThesis}.

\section{Acknowledgments}
\label{sec:acknowledgements}

Support provided by NSF PHY-1505048 and SES-1735579.

\bibliography{bib} 

\end{document}